\documentclass{article}
\usepackage{spconf,amsmath,graphicx,amssymb,setspace,url,multirow}
\usepackage{xcolor}

\title{The Dawn of Quantum Natural Language Processing}
\name{Riccardo Di Sipio$^1$, Jia-Hong Huang$^2$, Samuel Yen-Chi Chen$^3$, Stefano Mangini$^4$, Marcel Worring$^2$}
\address{$^1$Ceridian HCM Inc., $^2$University of Amsterdam, $^3$Brookhaven National Laboratory, $^4$University of Pavia}
\usepackage{ulem}

\newcommand{\softmax}{\texttt{softmax}~}
\newcommand{\ie}{\textit{i}.\textit{e}.~}
\newcommand{\eg}{\textit{e}.\textit{g}.~}

\begin{document}
%
\maketitle
\begin{abstract}
In this paper, we discuss the initial attempts at boosting understanding human language based on deep-learning models with quantum computing. We successfully train a quantum-enhanced Long Short-Term Memory network to perform the parts-of-speech tagging task via numerical simulations. Moreover, a quantum-enhanced Transformer is proposed to perform the sentiment analysis based on the existing dataset. 
\end{abstract}
\section{Introduction}
\label{sec:intro}

In recent times, large pre-trained neural network models have been successfully used to achieve state-of-the-art performance in computer vision (CV) \cite{krizhevsky2012imagenet,carion2020end,huang2019novel,he2016deep,huang2020query,yang2018novel,yang2018auto,liu2018synthesizing,huang2021deepopht,dosovitskiy2020image,braso2020learning,huang2017robustness,li2020recurrent,yi2020contextual,huang2021longer,guo2020learning,deng2021masked,hu2019silco,yang2020temporal,li2020smallbignet,yu2021frequency} and language processing (NLP) \cite{vaswani2017attention,devlin2018bert,peters2018deep,huang2021deep,huang2017robustnessMS,radford2019language,yang2021voice2series,yang2020characterizing,huang2019assessing,9413453,pan2020x,huang2017vqabq,pahwa2021medskip,fan2021beyond}. In particular, the field of NLP is seeing an exponential expansion in terms of real-life applications such as machine translation \cite{bahdanau2014neural}, document classification \cite{yang2016hierarchical}, interaction with a chatbot \cite{cahn2017chatbot}) but also network size as Transformer-based models seem to outperform long-standing architectures such as Long Short-Term Memory (LSTM) recurrent neural networks \cite{hochreiter1997long}. However, the main issue with this marvellous kind of neural networks is that the appalling size of parameters (in the order of hundreds of billions in the case of OpenAI’s GPT models \cite{brown2020language,huang2021gpt2mvs,huang2021contextualized,yang2021voice2series}) usually requires a training dataset of huge dimension such as the complete Wikipedia corpus in several languages, and a massive computer cluster to carry out the training. 
An estimation of the training cost in terms of the electricity bill is in the order of almost 5 million US dollars for GPT-3, and it would take more than 350 years if a single GPU was being used. Some see this as evidence that the entire field of artificial intelligence is becoming more and more undemocratic (a small company, let alone single person, simply can’t compete with tech giants), while others acknowledge that the progression resembles what happened in science in the past decades: it was only the concerted effort of national states over three decades that yielded the discovery of the Higgs boson at the CERN Large Hadron Collider \cite{Aad_2012,Chatrchyan_2012}. 

Finally, there is a subtle and rather philosophical issue concerning what these networks actually learn. Do these machine just replicate patterns seen in human documents, or do they have a broader understanding of what’s being said? A sentence such as “if you throw a metal ball against a window, you may break it” has an obvious meaning to anyone who has experience with metals and glass. This may not be the case for a neural network that has been only exposed to written text, with no connection to the physical world. On the other hand, associations between metals being hard and glass being brittle are likely to occur enough times in the text corpus to make this rather obvious even for the dumbest of the machines. A more detailed analysis of this point can be found in a famous piece by philosopher Douglas Hofstadter \cite{hofstadter2018shallowness}.

\begin{figure}[t!]
\begin{minipage}[b]{1.0\linewidth}
  \centering
  \scalebox{0.92}{
  \centerline{\includegraphics[width=8.5cm]{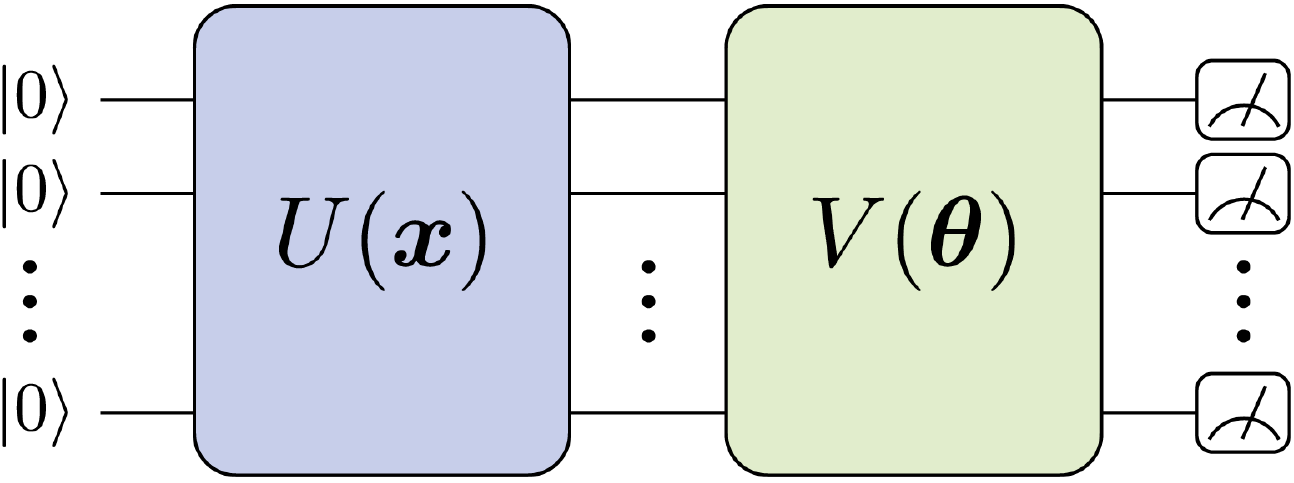}}
   }
\end{minipage}
\caption{\small{ Generic architecture for variational quantum circuits. $U(\mathbf{x})$ is the quantum routine for encoding the (classical) input data $\mathbf{x}$ into a quantum state, and $V(\boldsymbol{\theta})$ is the variational circuit block with tunable parameters $\boldsymbol{\theta}$.
A quantum measurement over some or all of the qubits follows.}}
\label{fig:vqc}
\end{figure}

On the other hand, the past years have witnessed the rise of quantum computing both in terms of hardware development~\cite{wendin2017quantum,mizuno2021research} and implementation of algorithms on such platforms~\cite{schuld2019evaluating,benedetti2021hardware}. In quantum computing~\cite{NielsenChuang}, computations are carried over the equivalent of bits called qubits, which notoriously can handle information in a non-binary state thanks to a property of quantum systems called superposition \cite{kazem2020effect}. A quantum circuit is a series of operations applied to qubits that change their state, \eg by changing their relative phase or angles in the quantum Hilbert space. Qubits can be represented geometrically with a so called Bloch sphere~\cite{kazem2020effect}, so an operation onto a qubit corresponds to a rotation of the quantum state vector in this virtual space. One key concept of quantum algorithms is that, because of the intrinsic nature of qubits~\cite{kazem2020effect}, certain calculations can be carried out with smaller complexity compared to the classical equivalent (this property is known as super-polynomial speedup~\cite{NielsenChuang, biamonteQuantumMachineLearning2017a, ben2020symmetries}). This is especially true in the case of chemistry, where quantum algorithms are used to predict electronic structures \cite{Whitfield_2011}, ground-state energies\cite{PhysRevX.8.031022}, and the spatial configuration of proteins \cite{fingerhuth2018quantum}. In fact, quantum computers are extremely efficient in cases where CPUs may take — literally — forever, such as in the simulation of the Ferredoxin molecule \cite{Wecker_2014}.


Practical applications are more likely to be a hybrid of classical and quantum operations since the currently available quantum computers are only small to intermediate scale. Under this hybrid paradigm, most of the calculations are carried out on classical computers, while the quantum devices are deployed to solve the computational tasks for which certain quantum advantages can be demonstrated. This hybrid approach is not too different from what has been done in the past decade with GPUs. Thus, the main idea behind Quantum Machine Learning (QML)~\cite{mangini2021quantum, chen2020variational} is to replace parts of a neural network (\eg linear layers) with a quantum counterpart. A comparison between different learning models is shown in Tab. \ref{tab:learning_models}.

Finally, as classical machine learning does not end with just neural network, also quantum machine learning find its strength in a variety of methods, such as quantum support vector machines~\cite{Havlicek2019SVM, heredge2021quantum}. While it is still too early to claim that quantum computing has taken over~\cite{huangPowerDataQuantum2021a}, there are some areas where it can give an advantage as for example in drug discovery~\cite{zinner2021quantum} or finance~\cite{egger2020quantum}. We argue that one field that so far has been poorly explored in QML is NLP.

\vspace{+5pt}
\noindent\textbf{Our contributions include:}
\begin{itemize}

    \item We are the first who successfully train a quantum-enhanced Long Short-Term Memory model to perform the parts-of-speech tagging task.
    \item We are the first to propose a quantum-enhanced Transformer model to perform the sentiment analysis task.
 
\end{itemize}


\section{Quantum Computing and Human Language}
\label{sec:quantum_language}
In early 2020, the British company called Cambridge Quantum Computing announced the introduction of a ``meaning-aware'' Quantum NLP model, \ie a mathematical theory able to splice the semantic information of words with the syntactic structure of a sentence~\cite{coecke2021make}. The bold claim is based on the observation that syntactic structures such as Noam Chomsky’s Context-free grammars or Categorical Compositional Distributional (DisCoCat) can be formulated in the framework of quantum physics~\cite{coecke2021make}. While the concept of a Hilbert space for NLP may seem far-fetched at first, an intuitive explanation goes as follows: NLP-as-we-know-it is rooted in classical statistics (for example, word embeddings are vectors in a $\mathbb{R}^d$ space~\cite{pennington2014glove}), but classical statistics itself has its own limits. The field of physics called quantum mechanics is described by the mathematics of quantum statistics, which extends classical statistics by representing objects with matrices of complex numbers. Even if we are not claiming that words are made of particles, it does happen by accident that the kind of statistics needed to describe human language is substantially the same of quantum physics. Coincidentally, we are also entering a time in which rudimentary and still quite noisy devices~\cite{Preskill2018NISQ} able to carry out computations based on qubits rather than digital bits of information are becoming more and more accessible to the wider public (including small businesses, start-ups and individuals). Thus, it makes sense in this decade to explore the possibility that quantum computing may give a boost to natural language processing.

\begin{table}[t!]
\begin{center}
\caption{\small{Comparison between different learning models.}}
\vspace{+0.1cm}
\begin{tabular}{ |p{0.25\linewidth}|p{0.25\linewidth}|p{0.25\linewidth}| }
 \hline
 {\bf Classical} & {\bf Quantum} & {\bf Hybrid} \\
 \hline\hline
bits & qubits & bits + qubits \\
\hline
DNN, GBDT & VQC & DNN + VQC \\
\hline
Broad applications & Quantum advantage & Broad applications with Quantum Advantage\\
\hline
Hardware-ready & Limited resources & Quantum hardware only where necessary \\
\hline
\end{tabular}
\label{tab:learning_models}
\end{center}
\end{table}

\section{Document Classification with Variational Quantum Circuits}
\label{sec:classification}
In the first example that we want to discuss, a deep neural network \cite{cer2018universal} is used to provide a sentence embedding. A final layer ("head") maps these embeddings to a probability vector of dimension ($n_{tokens}, n_{classes}$). 

In a classical machine learning setting, this is easily achievable by the means of a linear layer with \softmax activation. In the QML framework, the calculation is carried by first putting the input qubits into the initial state (\eg a string such as $010010$), then they are entangled between each other, rotated by arbitrary angles, and finally observed (“measured”). The goal of the training is to find the rotation angles mentioned above that optimize some cost function such as the mean squared error (MSE) between the output and some pre-determined labels. This kind of quantum circuit is known as Variational Quantum Circuit (VQC) \cite{mangini2021quantum, cerezoVariationalQuantumAlgorithms2021}. See Fig. \ref{fig:vqc} for VQC architecture visualization. However, a VQC cannot change the dimensionality of the input but only rotate the state of the qubits. Hence, the VQC is ``dressed'', \ie it is sandwiched between two classical linear layers to match the dimensionality. A classical layer ``squeezes'' the input to match the number of qubits, and another classical layer ``bloats'' the output to match the dimension of the hidden vectors. 

Preliminary results indicate that such a hybrid network can be successfully trained to perform classification. However, one has to keep in mind that the heavy lifting of this operation is performed by the sentence embedding deep network.

\section{Quantum-Enhanced Long Short-Term Memory Neural Network}
\label{sec:lstm}
As documents are usually presented as sequences of words, historically one of the most successful techniques to manipulate this kind of data has been the Recurrent Neural Network (RNN) architecture, and in particular a variant called Long Short-Term Memory (LSTM) \cite{hochreiter1997long}. The ``trick'' that makes these networks so popular in the analysis of sequential data is a combination of ``memory'' and ``statefulness'' that help with identifying which components of the input are relevant to compute the output. While the mathematics is quite thick as we will see later on, it suffices to say for now that LSTMs allowed machines to perform translations, classification and intent detection with state-of-the-art accuracy until the advent of Transformer networks \cite{vaswani2017attention}. Still, it is interesting at least from an educational point of view to dig into LSTMs to see what good quantum computing may bring to the field. For a more thorough discussion, refer to ``Quantum Long Short-Term Memory'' \cite{chen2020quantum} and ``Recurrent Quantum Neural Network'' \cite{bausch2020recurrent}.

\begin{figure}[htb]
\begin{minipage}[b]{1.0\linewidth}
  \centering
  \centerline{\includegraphics[width=8.5cm]{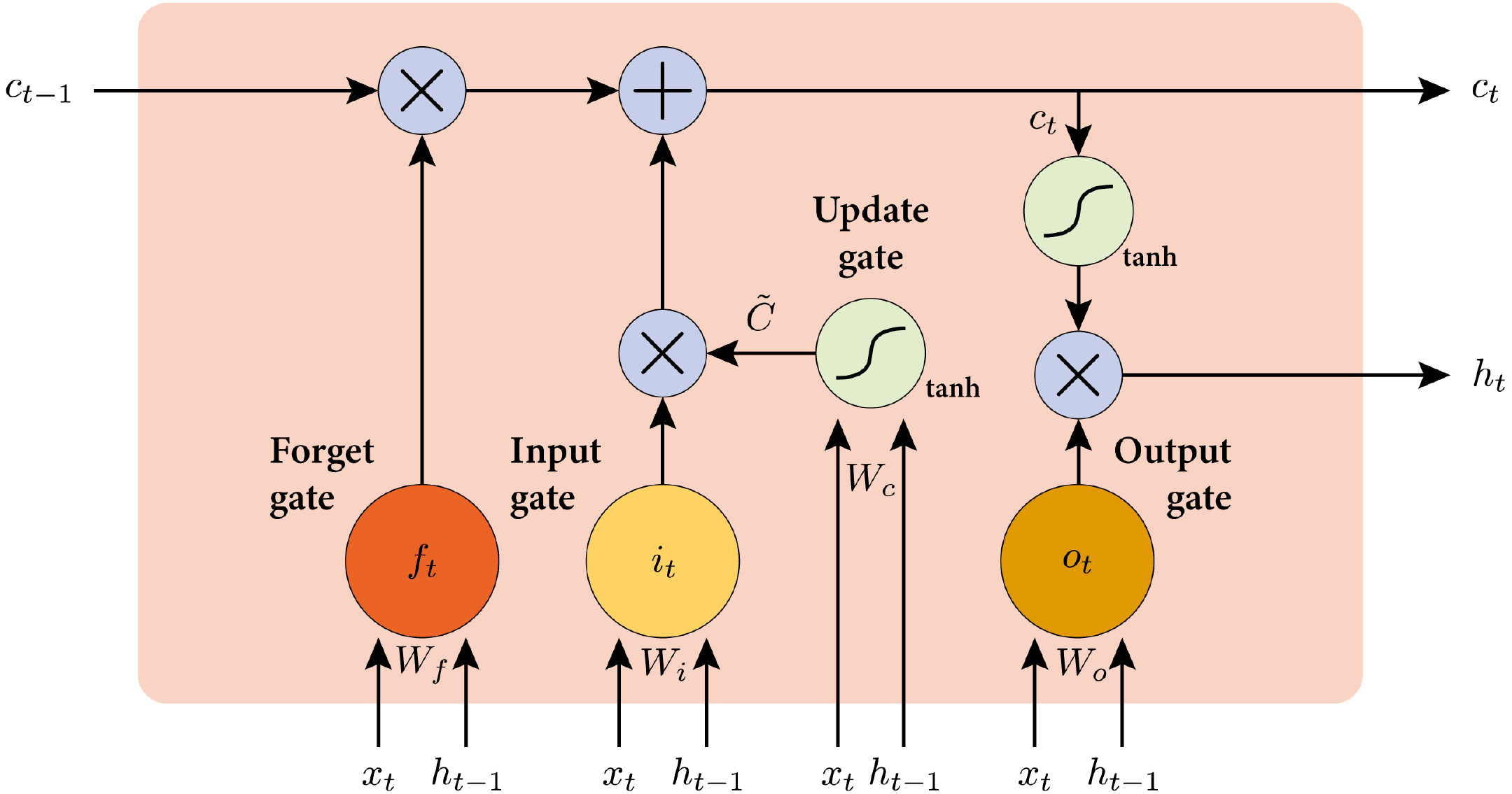}}
\end{minipage}
\caption{\small{Graphical representation of the information flow and operations carried out inside a LSTM neural network layer. In the quantum-enhanced LSTM (QLSTM), each of the classical operation $W_{f}$, $W_{i}$, $W_{C}$ and $W_{o}$ is replaced by a hybrid quantum-classical component which includes a VQC sandwiched between classical layers.}}
\label{fig:lstm}
\end{figure}

To begin with, let’s review the inner workings of a LSTM. We assume that the input is composed of a sequence of $t$ time-steps (\eg words), each represented by a $N$-dimensional feature vector (in practical applications $N$ can be large, \eg 512). Also, the network stores a hidden array of vectors $h$ and a state vector $c$ that are updated for each element of the input sequence. For example, if we are only interested in a summary of the sequence (\eg in a sentiment analysis), the last element of the array $h$ will be returned. Instead, if we are interested in a representation of each element (\eg to assign to each word a part-of-speech tag such as noun, verb, etc), we want to have access to every element of $h$. The calculation can be summarized in the formulas below:

\begin{eqnarray}
f_t & = & \sigma(W_f \cdot v_t + b_f) \\
i_t & = & \sigma(W_i \cdot v_t + b_i) \\
\tilde{C} & = & \tanh (W_C \cdot v_t + b_C) \\
c_t & = & f_t \odot c_{t-1} + i_t \odot \tilde{C}\\
o_t & = & \sigma(W_o \cdot v_t + b_o) \\
h_t & = & o_t \odot \tanh (c_t)
\end{eqnarray}

where $v_t$ is a concatenation of the input element at step $t$ and the hidden state at step $t-1$, \ie $v_t = [h_{t-1}, x_t]$. In the machine learning lingo, borrowed from electric circuit analysis, $f_t$ is called forget gate, $i_t$ is the input gate, $c_t$ is the update gate and $o_t$ is the output gate. The matrices $W_f$, $W_i$, $W_c$ and $W_o$ and the bias vectors $b_f$, $b_i$, $b_c$ and $b_o$ are the parameters that have be learned during the supervised training and implement the part of the calculation called linear dense layer that we want to replace with the quantum equivalent. As is often the case, a non-linearity is introduced by the application of sigmoid ($\sigma$) and hyperbolic tangent ($\tanh$) functions to the output of these four dense layers, which effect is to determine whether a part of the input has to be considered (values close to $1$) or ignored (values close to $0$). Fig. \ref{fig:lstm} shows a graphical representation of the information flow and operations carried out inside a LSTM.

Notably, the linear layers of LSTM change the dimensionality of the input tensor, something that is not possible with qubit rotations. As described in Sec. \ref{sec:classification}, the VQC layers have to be ``dressed'', \ie sandwiched between two classical linear layers to match the dimensionality. This is applied to $f_t$, $i_t$, $c_t$ and $o_t$ separately. Thus, overall there is still a sizeable number of classical parameters to be learned during the training.

\begin{figure}[htb]
\begin{minipage}[b]{1.0\linewidth}
  \centering
  \centerline{\includegraphics[width=8.5cm]{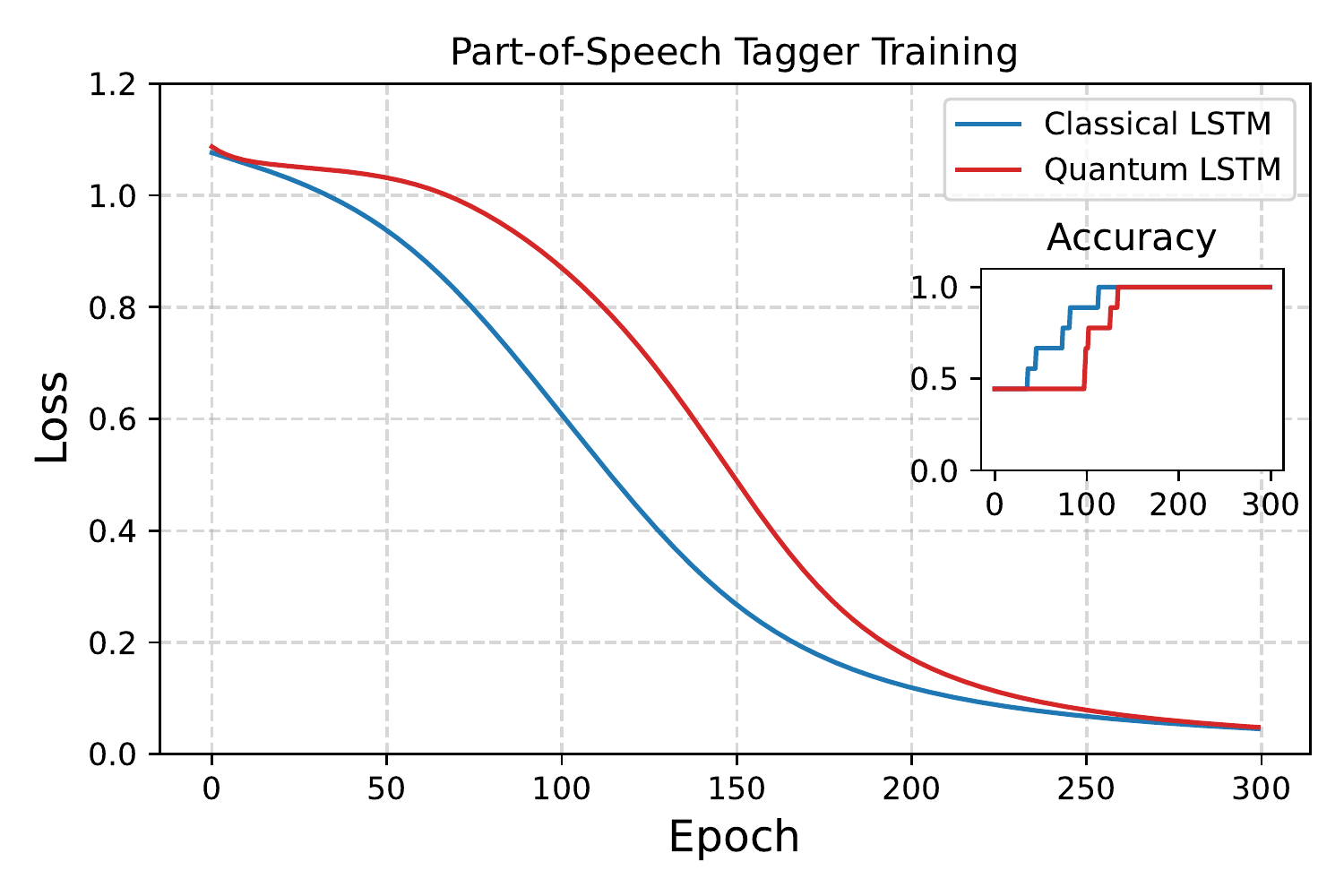}}
\end{minipage}
\caption{\small{Cross-entropy loss and multi-class accuracy as a function of the training epoch for the classical and quantum LSTM Part-of-Speech taggers.}}
\label{fig:qlstm}
\end{figure}

To test the setting described above, an intuitive but not trivial example is provided: a {\sl part-of-speech tagging} task. In an example, two sentences (``The dog ate the apple'' and ``Everybody read that book'') have been annotated with POS tags. For example, the first sentence is [``DET'', ``NN'', ``V'', ``DET'', ``NN'']. The LSTM will output the hidden array of vectors $[h_0, h_1, h_2, h_3, h_4]$, one for each word. A dense layer ``head'' with \softmax activation is attached to the LSTM’s outputs to calculate the probability that each word may be a determinant, noun or verb. We trained the two networks (classical and quantum LSTM) for $300$ epochs each. The parameters used to define the quantum network are described in Tab. \ref{tab:qlstm}. As shown in Fig. \ref{fig:qlstm}, the cross-entropy loss decreases as a function of the training epoch, and after 150 epochs both networks are able to tag correctly the two sentences. Due to the complexity of the simulation of the quantum circuit, it took approximately $10$ minutes to finish the training, to be compared to a mere 10 seconds for the classical case. Also, it seems that the quantum LSTM needed to be trained for longer epochs to achieve 100\% accuracy. At the end of the $300$ epochs, the loss of the classical network is $0.053$ while that of the quantum network is $0.056$. It is worth noticing that the quantum LSTM uses less than half the parameters of the purely classical one ($199$ vs. $477$), while retaining the same overall performances. The open-source code is available at: \protect\url{https://github.com/rdisipio/qlstm}.

\begin{table}[htb]
\begin{center}
\caption{\small{Parameters used to define the quantum-enhanced LSTM network.}}
\vspace{+0.1cm}
\begin{tabular}{ |l|c| } 
 \hline
 Parameter & Value \\ 
 \hline\hline
 epochs & 300 \\
 vocab size & 5 \\
 number of tags & 3 \\
 embedding dim & 8 \\
 hidden dim & 6 \\
 no. of quantum layers & 1 \\
 no. of qubits in VQC & 4\\
 \hline
 Total number of weights & 199 \\
 \hline
\end{tabular}
\label{tab:qlstm}
\end{center}
\vspace{-0.6cm}
\end{table}

\section{Toward a Quantum Transformer}
\label{sec:pagestyle}

As mentioned in the introduction, the Transformer architecture \cite{vaswani2017attention} revolutionized the analysis of sequential data, and in particular that of human-written documents. Transformers are a neural network architecture that is optimized to analyze sequential data on highly-parallel devices such as GPUs and TPUs. Differently from recurrent networks, Transformers do not have ``memory'' but are still able to perform the trick by a combination of position-dependent embeddings (\ie words embeddings are supplemented by another set of vectors that depend on the position the word has in the sentence, and on the index of the embedding dimension) and attention (\ie figuring out which parts of the inputs are relevant to calculate the output). At the core of any Transformer sits the so-called Multi-Headed Attention \cite{vaswani2017attention}. The idea is to apply three different linear transformations $W_Q$, $W_K$ and $W_V$ to each element of the input sequence to transform each word embedding into some other internal representation states called Query ($Q$), Key ($K$), and Value ($V$) \cite{vaswani2017attention}. These states are then passed to the function that calculates the attention weights, which is simply defined as:

\begin{equation}
    Attention(Q, K, V)= softmax\left(\frac{QK^T}{\sqrt{d_k}}\right)V
\end{equation}

To promote the Transformer from the classical to quantum real, one can replace the linear transformations $W_Q$, $W_K$ and $W_V$ with VQCs \cite{chen2020variational}. A preliminary experiment for sentiment analysis made use of the IMDB dataset~\cite{maas-EtAl:2011:ACL-HLT2011} for binary classification. We used PennyLane \cite{bergholm2020pennylane} version 0.13.0 and its plugin to perform simulations of quantum processes with the Qulacs library \cite{suzuki2021qulacs}, which is a VQC simulator that runs on a GPU. 
The parameters of the hybrid network are described in Tab. \ref{tab:qtransformer}. It took about $100$ hours to train the classifier for a single epoch. It is clear that unless a more sophisticated training method is devised, model development and parameter tuning is prohibitive with the hardware currently available to the general public. The open-source code is available at \protect{\url{https://github.com/rdisipio/qtransformer}}.

\begin{table}[htb]
\begin{center}
\caption{\small{Parameters used to define the proposed quantum-enhanced transformer network.}}
\vspace{+0.1cm}
\begin{tabular}{ |l|c| } 
 \hline
 Parameter & Value \\ 
 \hline\hline
 batch size & 32 \\ 
 epochs & 1 \\ 
 vocab size & 50,000 \\
 embedding dim & 8 \\
 max seq length & 64 \\
 feed-forward net dim & 8 \\
 drop-out rate & 0.1 \\
 no. of transformer blocks & 1 \\
 no. of transformer heads & 2 \\
 no. of quantum layers & 1 \\
 no. of qubits in transformer blocks & 2 \\
 no. of qubits in feed-forward net dim & 2\\
 \hline
  Total number of weights & $\mathcal{O}$(150,000)  \\
\hline
\end{tabular}
\label{tab:qtransformer}
\end{center}
\vspace{-0.6cm}
\end{table}

\section{Conclusions}
\label{sec:conclusions}
It is a legitimate question whether it would be possible to devise a fully-quantum Transformer, \ie a quantum circuit acting upon qubits in a fashion similar to a combination self-attention and positional embedding. While we cannot offer a definite answer here, we argue that most of the building blocks of the basic Transformer have a corresponding quantum operation, as arithmetic operations and exponentiation can be implemented as quantum circuits~\cite{QuantumArithmetics}, or that they can be cast in form of variational quantum circuits. It is the sheer size of the computation that is likely beyond the limits of current quantum computers, although we foresee that this situation may change rapidly over the course of this decade. The end of this paper is a good place to circle back to the original question of what a Transformer actually is, and why it is so successful at interpreting human language. While a number of interpretations have been given in terms of computations on graphs~\cite{NEURIPS2019_9d63484a, cai2020graph}, in due course quantum computing may offer alternative ways to apply attention to sequential data, or provide some completely different conceptual framework altogether~\cite{lorenz2021qnlp}. Only time will tell.  



\begin{spacing}{0.6}
\footnotesize
\bibliographystyle{IEEEbib}
\bibliography{refs}
\end{spacing}
\end{document}